# Combinatorial Optimization by Learning and Simulation of Bayesian Networks

Pedro Larrañaga, Ramon Etxeberria, José A. Lozano and José M. Peña
Intelligent Systems Group
Dept. of Computer Science and Artificial Intelligence
University of the Basque Country
E-20080 Donostia–San Sebastián, Spain
ccplamup@si.ehu.es, ramon@edunet.com, lozano@si.ehu.es, ccbpepaj@si.ehu.es

## Abstract

This paper shows how the Bayesian network paradigm can be used in order to solve combinatorial optimization problems. To do it some methods of structure learning from data and simulation of Bayesian networks are inserted inside Estimation of Distribution Algorithms (EDA). EDA are a new tool for evolutionary computation in which populations of individuals are created by estimation and simulation of the joint probability distribution of the selected individuals. We propose new approaches to EDA for combinatorial optimization based on the theory of probabilistic graphical models. Experimental results are also presented.

## 1 Introduction

Roughly speaking, search strategies can be classified as complete or heuristic strategies. The underlying idea in the complete search is the systematic examination of all the possible points of the search space. On the other hand, heuristic algorithms can be classified as deterministic or non-deterministic. In deterministic heuristic strategies the same solution is always achieved under the same conditions. Non-deterministic search is motivated by trying to avoid getting stuck in local maximum. Randomness is used to escape from local maximum and, due to its stochasticity, different runs might lead us to achieve different solutions under the same conditions. While some of the stochastic heuristic strategies (e.g., simulated annealing) store only one solution in each iteration of the algorithm, in other approaches –evolutionary computation– the search is based on a population of individuals (each of which represent a point of the search space). This population evolves as the algorithm proceeds toward more promising zones of the space of solutions. Examples of evolutionary computation are genetic algorithms, evolutionary strategies, evolutionary programming and genetic programming.

The behavior of the addressed evolutionary computation algorithms depends on several parameters associated with them (operators of crossing and mutation, probabilities of crossing and mutation, size of the population, rate of generational reproduction, number of generations, ...). If the researcher does not have experience in the resolution of a concrete optimization problem by means of this type of approach, the choice of the suitable values for all the parameters is itself converted into an optimization problem. This reason, together with the fact that the prediction of the movements of the populations in the search space is extremely difficult, has motivated the birth of a type of algorithms denominated Estimation of Distribution Algorithms (EDA) (Mühlenbein and Paaß 1996).

In EDA there are neither crossover nor mutation operators and, instead, the new population of individuals is sampled from the probability distribution, which is estimated from the database containing only selected individuals from the previous generation. Whereas in the heuristics coming from evolutionary computation, the interrelations between the different variables representing the individuals are kept in mind implicitly –building block hypothesis–, in EDA, the interrelations are expressed in an explicit way through the joint probability distribution associated with the individuals selected in each iteration. In fact, the estimation of the joint probability distribution associated with the data base containing the selected individuals constitutes the bottleneck of this new heuristic.

The fundamental objective of this work is to propose new EDA for combinatorial optimization problems. These new algorithms are based on three different approaches to the structure learning of Bayesian networks: (i) detecting conditional independencies, (ii) penalized maximum likelihood, and (iii) Bayesian approach.

The outline of this work is as follows. Section 2 introduces the EDA approach, classifying the different methods found in the literature according to the complexity of the probabilistic model used. In Section 3 a brief presentation of the different approaches to structure learning and simulation of Bayesian networks is done. In Section 4 we present new approaches to combinatorial optimization based on induction and simulation of Bayesian networks, while in Section 5 we show some experimental results. Section 6 picks up the conclusions of our work and some notes about possible future lines of research.

## 2 Estimation of Distribution Algorithms

### 2.1 Introduction

The poor behavior of genetic algorithms in some problems –deceptive problems– in which the designed operators of crossing and mutation do not guarantee the preservation of the building block hypothesis have led to the development of other type of algorithms.

These referred algorithms are known as Estimation of Distribution Algorithms (EDA). EDA are population-based search algorithms that use probabilistic modeling of promising solutions in combination with the simulation of the induced models to guide their search.

The underlying idea of EDA is introduced starting from a problem that arises in the supervised clasification and that is known as feature subset selection (FSS) –see Inza et al. (1999)–. Given a file of cases with information on $n$ predictive variables, $X_1, X_2, \ldots, X_n$ and the class variable $C$ to which the case belongs, the problem consists of selecting a subset of variables that will induce a classifier with the highest predictive capacity in a test set. The cardinality of the search space is $2^n$.

Figure 1 shows an schematic of the EDA approach. In the first step $N$ individuals are generated at random, for example, based on an uniform distribution on each variable. These $N$ individuals constitute the initial population, $D_0$, and each of them is evaluated. In a second step, a number $Se$ ($Se < N$) of individuals are selected (usually those with the higher objective function value). Next, the induction of the $n$–dimensional probabilistic model that best reflects the interdependences between the $n$ variables is carried out. In a fourth step, $N$ new individuals (the new population) are obtained by means of the simulation of the probability distribution learnt in the previous step. Steps 2, 3 and 4 are repeated until a stopping condition is verified.

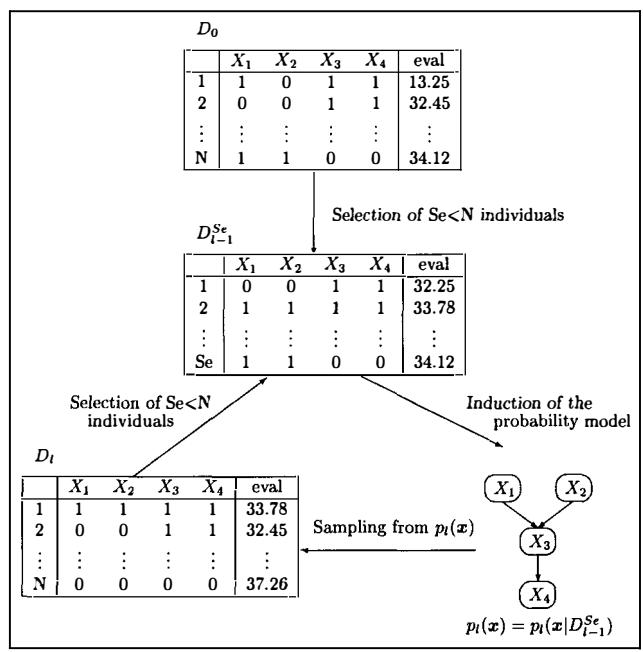

Figure 1: Illustration of the EDA approach to optimization

Let us introduce a general notation that will be used to express a pseudocode of EDA in combinatorial optimization problems.

We use $X_i, i = 1, \ldots, n$, to represent a random variable. A possible instantiation of $X_i$ is denoted $x_i$. $p(X_i = x_i)$ (or simply $p(x_i)$) represents the mass probability for the variable $X_i$ over the point $x_i$. Similarly, we use $\boldsymbol{X} = (X_1, \ldots, X_n)$ to represent an $n$–dimensional random variable and $\boldsymbol{x} = (x_1, \ldots, x_n)$ to represent one of its possible instantiations. The joint probability mass of $\boldsymbol{X}$ is denoted $p(\boldsymbol{X} = \boldsymbol{x})$ (or simply $p(\boldsymbol{x})$). The conditional mass probability of the variable $X_i$ given the value $x_j$ of the variable $X_j$ is represented as $p(X_i = x_i | X_j = x_j)$ (or simply by $p(x_i | x_j)$). We will use $D$ to represent a data set, i.e., a set of $N$ instantiations of the variables $(X_1, \ldots, X_n)$.

Connecting this notation with the one we need to express the pseudocode of EDA –see Figure 2–, $\boldsymbol{x} = (x_1, \ldots, x_n)$ will denote individuals of $n$ genes, and $D_l$ will denote the population of $N$ individuals in generation $l$. Similarly, $D_l^{Se}$ will represent the population of the selected $Se$ individuals from $D_l$. In EDA our interest will be to estimate $p(\boldsymbol{x} \mid D^{Se})$, that is, the joint mass probability over one individual $\boldsymbol{x}$ being among the selected individuals. This joint probability mass must be estimated in each generation. We denote the joint probability mass of the $l$-th generation as $p_l(\boldsymbol{x}) = p_l(\boldsymbol{x} \mid D_{l-1}^{Se})$.

```
EDA
  D_0 ← Generate N individuals (the initial population) randomly
  Repeat for l = 1, 2, ... until a stopping criterion is met
    D_{l-1}^{Se} ← Select Se ≤ N individuals from D_{l-1} according to a
    selection method
    p_l(x) = p(x|D_{l-1}^{Se}) ← Estimate the probability distribution
    of an individual being among the selected individuals
    D_l ← Sample N individuals (the new population) from p_l(x)
```

Figure 2: Pseudocode for EDA approach

The aim of this new paradigm is to try to detect the interdependences between the variables that represent one point in the search space. The basic idea consists of inducing probabilistic models from the best individuals found. Afterwards the models are sampled generating new solutions, which are used to update the model. This cycle is repeated until a stopping criterion is verified. The main problem of EDA lies on how the probability distribution $p_l(x)$ is estimated. Obviously, the computation of all the parameters needed to specify the probability model is impractical. This has led to several approximations where the probability distribution is assumed to factorize according to a probability model.

## 2.2 Proposed EDA in combinatorial optimization

In this subsection a revision of the EDA approaches found in the literature for combinatorial optimization problems will be carried out. See Larrañaga et al. (1999a, 1999b) for a more detailed revision.

### 2.2.1 Without interdependencies

In all the works belonging to this category it is assumed that the $n$–dimensional joint probability distribution factorizes as a product of $n$ univariate and independent probability distributions. That is $p_l(x) = \prod_{i=1}^{n} p_l(x_i)$. Obviously this assumption is very far from what happens in a difficult optimization problem where the interdependencies between the variables will exist in some degree.

Approaches in this category include the following ones: Bit-Based Simulated Crossover (BSC) (Syswerda 1993), Population-Based Incremental Learning (PBIL) (Baluja 1994), the compact Genetic Algorithm (cGA) (Harik et al. 1998), and the Univariate Marginal Distribution Algorithm (UMDA) (Mühlenbein 1998).

In the Univariate Marginal Distribution Algorithm (UMDA) (Mühlenbein 1998) $p_l(x_i)$ is estimated from the relative marginal frequencies of the $i$-th variable of the selected individuals $D_{l-1}^{Se}$.

### 2.2.2 Pairwise dependencies

The estimation of the joint probability distribution can also be done quickly only taking dependencies between pairs of variables into account –see Figure 3. On the other hand, where in the algorithms of the previous subsection only a learning of the parameters was carried out –the structure of the model remained fixed– in this subsection, the parametric learning is extended to structural.

In De Bonet et al. (1997) a greedy algorithm called MIMIC (Mutual Information Maximization for Input Clustering) is developed. MIMIC searches for the best permutation between the variables in order to find the probability distribution belonging to the class $\mathcal{P}_\pi(x)$, where:

$$\mathcal{P}_\pi(x) = \{p_\pi(x) \mid p_\pi(x) =$$
$$p(x_{i_1} \mid x_{i_2}) \cdot p(x_{i_2} \mid x_{i_3}) \cdots p(x_{i_{n-1}} \mid x_{i_n}) \cdot p(x_{i_n})\}, \quad (1)$$

that is closest with respect to the Kullback-Leibler distance to the set of selected points. It is easy to prove that the agreement between two probability distributions, $p(x)$ and $p_\pi(x)$, being $p_\pi(x) \in \mathcal{P}_\pi(x)$, measured by its Kullback-Leibler divergence is a function of $H_\pi(x) = h(X_{i_n}) + \sum_{j=1}^{n-1} h(X_{i_j} \mid X_{i_{j+1}})$, where $h(X_{i_j} \mid X_{i_{j+1}})$ denotes the mutual information between the variables $X_{i_j}$ and $X_{i_{j+1}}$, and $h(X_{i_n})$ is the entropy of the variable $X_{i_n}$. Applying this result, the problem of searching for the best $p_\pi(x) \in \mathcal{P}_\pi(x)$ is equivalent to searching for the permutation, $\pi^*$, of the $n$ variables minimizing $H_\pi(x)$. To find $\pi^*$ the authors propose a straightforward greedy algorithm avoiding searching through all $n!$ permutations. The idea is to select $X_{i_n}$ as the variable with the smallest estimated entropy and, afterwards, at every step, to pick up the variable –from the set of variables not chosen so far– whose average conditional entropy with respect to the previous is the smallest.

Another approaches in this group are the proposed by Baluja and Davies (1997) and the BMDA (Bivariate Marginal Distribution Algorithm) of Pelikan and Mühlenbein (1999).

### 2.2.3 Multiply interdependencies

Several approaches to EDA have been proposed in the literature in which the factorization of the joint probability distribution requires of statistics of order greater than two –see Figure 4.

In the work of Mühlenbein et al. (1999) the FDA (Factorized Distribution Algorithm) is introduced. This algorithm applies to additively decomposed functions for which, using the running intersection property, a factorization of the mass-probability based on residuals and separators is obtained.

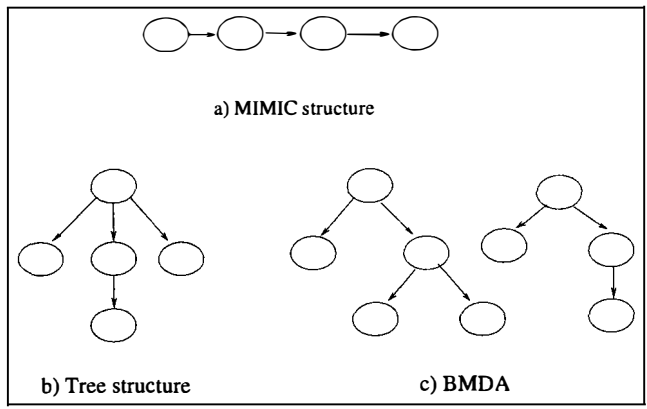

Figure 3: Graphical representation of proposed EDA in combinatorial optimization with pairwise dependencies (MIMIC, tree structure, BMDA)

The work of Etxeberria and Larrañaga (1999) is the first published paper in which a factorization of the joint probability distribution encoded by a Bayesian network is learnt from the database containing the selected individuals in each generation. The algorithm developed is called EBNA (Estimation of Bayesian Networks Algorithm) and makes use of the BIC approach applied to the Bayesian quality together with greedy algorithms to perform the search in the space of models.

In BOA (Bayesian Optimization Algorithm), proposed by Pelikan et al. (1999), a Bayesian metric, BDe, to measure the goodness of every structure found and a greedy search procedure are used. The search begins at each generation from scratch.

Mühlenbein and Mahnig (1999) introduce the LFDA (Learning Factorized Distribution Algorithm) which essentially follows the same approach as in EBNA.

Harik (1999) presents an algorithm –Extend compact Genetic Algorithm (EcGA)– whose basic idea consists of factorizing the joint probability distribution in a product of marginal distributions of variable size.

## 3 Structure learning and simulation in Bayesian networks

### 3.1 Introduction

Bayesian networks have been surrounded by a growing interest in the recent years, shown by the large number of dedicated books and a wide range of theoretical and practical publications in this field. Textbooks include the following ones: Pearl (1988), Jensen (1996) and Castillo et al. (1997). More recently Cowell et al. (1999) provides an excellent compilation material covering recent advances in the field.

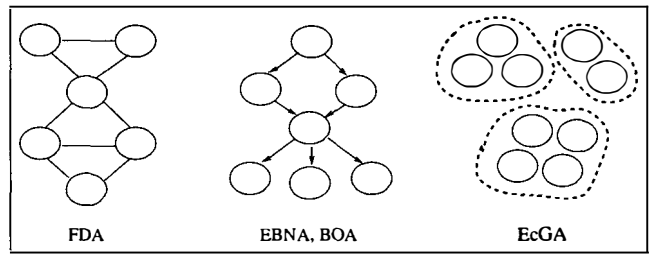

Figure 4: Graphical representation of proposed EDA in combinatorial optimization with multiply dependencies (FDA, EBNA, BOA and EcGA)

Let $X = (X_1, \ldots, X_n)$ be a set of random variables. We will use $x_i$ to denote a value of $X_i$, the $i$-th component of $X$, and $y = (x_i)_{X_i \in Y}$ to denote a value of $Y \subseteq X$. A probabilistic graphical model for $X$ is a graphical factorization of the joint generalized probability density function, $\rho(X = x)$ (or simply $\rho(x)$). The representation consists of two components: a structure and a set of local generalized probability densities. The structure $S$ for $X$ is a directed acyclic graph (DAG) that represents a set of conditional independences[1] assertions on the variables on $X$.

The structure $S$ for $X$ represents the assertions that $X_i$ and its non descendents are independent given $Pa_i^S$ [2], $i = 2, \ldots, n$. Thus, the factorization is as follows:

$$\rho(x) = \rho(x_1, \ldots, x_n) = \prod_{i=1}^{n} \rho(x_i \mid pa_i^S). \quad (2)$$

The local generalized probability densities associated with the probabilistic graphical model are precisely those in the previous equation.

In this paper, we assume that the local generalized probability densities depend on a finite set of parameters $\theta_S \in \Theta_S$. Thus, we rewrite the previous equation as follows:

$$\rho(x \mid \theta_S) = \prod_{i=1}^{n} \rho(x_i \mid pa_i^S, \theta_i) \quad (3)$$

---

[1] Given $Y, Z, W$ three disjoints sets of variables, we said that $Y$ is conditionally independent of $Z$ given $W$ if for any $y, z, w$ we have $\rho(y \mid z, w) = \rho(y \mid w)$.

[2] $Pa_i^S$ represents the set of parents –variables from which an arrow is coming out– of the variable $X_i$ in the probabilistic graphical model with structure given by $S$.

where $\boldsymbol{\theta}_S = (\boldsymbol{\theta}_1, \ldots, \boldsymbol{\theta}_n)$. Taking both components of the probabilistic graphical model into account, this will be represented by $M = (S, \boldsymbol{\theta}_S)$.

In the particular case of every variable $X_i \in \boldsymbol{X}$ being discrete, the probabilistic graphical model will be called *Bayesian network*. If the variable $X_i$ has $r_i$ possible values, $x_i^1, \ldots, x_i^{r_i}$, the local distribution, $p(x_i \mid \boldsymbol{pa}_i^{j,S}, \boldsymbol{\theta}_i)$ is an unrestricted discrete distribution:

$$p(x_i^k \mid \boldsymbol{pa}_i^{j,S}, \boldsymbol{\theta}_i) = \theta_{x_i^k \mid \boldsymbol{pa}_i^j} \equiv \theta_{ijk} \qquad (4)$$

where $\boldsymbol{pa}_i^{1,S}, \ldots, \boldsymbol{pa}_i^{q_i,S}$ denotes the values of $\boldsymbol{Pa}_i^S$, the set of parents of the variable $X_i$ in the structure $S$. The number $q_i$ denotes the number of possible different instantiations of the parent variables of $X_i$. Thus, $q_i = \prod_{X_g \in \boldsymbol{Pa}_i} r_g$. The local parameters are given by $\boldsymbol{\theta}_i = ((\theta_{ijk})_{k=1}^{r_i})_{j=1}^{q_i}$. In others words, the parameter $\theta_{ijk}$ represent the conditional probability of variable $X_i$ being in its $k$-th value, given that the set of its parent variables is in its $j$-th value. We assume that each parameter $\theta_{ijk}$ is greater than zero.

### 3.2 Structure learning

Once the Bayesian network is built, it constitutes an efficient device to perform probabilistic inference. Nevertheless, the problem of building such a network remains. The structure and conditional probabilities necessary for characterizing the Bayesian network can be provided either externally by experts –time consuming and subject to mistakes– or by automatic learning from a database of cases. On the other hand, the learning task can be separated into two subtasks: *structure learning*, that is, to identify the topology of the Bayesian network, and *parametric learning*, the numerical parameters (conditional probabilities) for a given network topology.

The different approaches to Bayesian network model induction can be classified from a double perspective. Firstly, we take the complexity of the learnt model into account (tree, polytree or multiply connected). Secondly, we can consider the nature of the modeling (detecting conditional independencies versus score+search).

The reader can consult some good reviews about model induction in Bayesian networks in Heckerman (1995) and Buntine (1996).

### 3.3 Simulation

A good number of methods for the simulation of Bayesian networks have been developed during the last years. In this paper, we will use the method *Probabilistic Logic Sampling* (PLS) proposed by Henrion (1988).

In this method the instantiations are generated one variable at a time in a forward way, that is, a variable is sampled after all its parents have already been sampled. Thus, variables must be ordered in such a way that the values for $\boldsymbol{Pa}_i$ must be assigned before $X_i$ is sampled. An ordering of the variables satisfying such property is called an ancestral ordering. Once the values of $\boldsymbol{Pa}_i$ have been assigned, we simulate a value for $X_i$ using the distribution $p(x_{\pi(i)}|\boldsymbol{pa}_i)$. Figure 5 shows a pseudocode of the method.

```
PLS
    Find an ancestral ordering, π, of the nodes
    in the Bayesian network
    For j = 1, 2, ..., N
    For i = 1, 2, ..., n
        x_π(i) ← generate a value from p(x_π(i)|pa_i)
```

Figure 5: Pseudocode for the Probabilistic Logic Sampling method

## 4 New approaches to combinatorial optimization based on learning and simulation of Bayesian networks

In this section three different approaches to combinatorial optimization based on the learning and simulation of Bayesian networks will be introduced.

### 4.1 Detecting conditional independencies $EBNA_{PC}$

Among the many methods for recovering Bayesian networks by means of detecting conditional independencies, we will use the PC algorithm (Spirtes et al. 1991) in order to obtain the structure which best fits $p(\boldsymbol{x}|D_{l-1}^{Se})$. The PC algorithm starts by forming the complete undirected graph and then it tries to "thin" it. First, edges with zero order conditional independence relations are removed, then edges with first order conditional independence relations, and so on. In order to check the conditional independencies Chi square tests ($\alpha = 0.01$) will be performed.

### 4.2 Penalized maximum likelihood $EBNA_{BIC}$

This approach corresponds to that developed by Etxeberria and Larrañaga (1999). In their paper, the score used to evaluate the goodness of each structure found during the search is the penalized maximum likelihood. In particular, they propose the use of the Bayesian Information Criterion (BIC) (Schwarz 1978).

This means that to evaluate a Bayesian network structure $S$, from a database $D$ containing $N$ cases, its cor-responding BIC score, denoted $BIC(S, D)$, is as follows:

$$BIC(S, D) = \sum_{i=1}^{n} \sum_{j=1}^{q_i} \sum_{k=1}^{r_i} N_{ijk} \log \frac{N_{ijk}}{N_{ij}} - \frac{\log N}{2} \sum_{i=1}^{n} (r_i - 1) q_i \quad (5)$$

where $N_{ijk}$ and $N_{ij}$ and $q_i$ are defined as above.

The quantitative part of the Bayesian network, that is, the parameters of the local probability distributions, are calculated every generation using their expected values as obtained by Cooper and Herskovits (1992):

$$E[\theta_{ijk} \mid S, D] = \frac{N_{ijk} + 1}{N_{ij} + r_i}. \quad (6)$$

---

$EBNA_{BIC}$
  $M_0 \leftarrow (S_0, \boldsymbol{\theta}_0)$
  $D_0 \leftarrow$ Sample $N$ individuals from $M_0$
  For $l = 1, 2, \ldots$ until a stop criterion is met
    $D_{l-1}^{Se} \leftarrow$ Select $Se$ individuals from $D_{l-1}$
    $S_l^* \leftarrow$ Find the structure which maximizes
    $BIC(S_l, D_{l-1}^{Se})$
    $\boldsymbol{\theta}^l \leftarrow$ Calculate $\{\theta_{ijk}^l = \frac{N_{ijk}^l + 1}{N_{ij}^l + r_i}\}$ using $D_{l-1}^{Se}$
    as the data set
    $M_l \leftarrow (S_l^*, \boldsymbol{\theta}^l)$
    $D_l \leftarrow$ Sample $N$ individuals from $M_l$ using PLS

---

Figure 6: Pseudocode for $EBNA_{BIC}$ algorithm

Unfortunately, finding the best model requires searching through all possible structures, which has been proved to be NP-hard (Chickering et al. 1994). Although promising results have been obtained using global search techniques (Larrañaga et al. 1996a, 1996b and Etxeberria et al. 1997a, 1997b, Myers et al. 1999), their computation cost makes them unfeasible for our problem. We need to find a model good enough as quickly as possible, so a simple algorithm which returns a good structure, even if it is not optimal, is preferred. An interesting algorithm with these characteristics is the Algorithm B (Buntine 1991). The Algorithm B is a greedy search heuristic which starts with an arc-less structure and, at each step it adds the arc with the maximum improvement in the BIC approximation (or whatever measure is used). The algorithm stops when adding an arc would not increase the scoring measure used.

Another possibility to find good models in a fast way consists in the use of local search strategies. Starting with a given structure, at every step the addition or deletion of an arc with the maximum increase in the scoring measure is performed. Local search strategies stop when no modification of the structure improves the scoring measure. The main drawback of local search strategies is that they heavily depend on the initial structure. Hovewer, Chickering et al. (1995) reported that local search strategies perform quite well when the initial structure is reasonably good. Moreover, if we base our search on the assumption that $p(x|D_l^{Se})$ will not differ very much from $p(x|D_{l-1}^{Se})$, the model found in the previous generation could be used as the initial one for the current generation.

In the initial model –see Figure 6–, $M_0$, the structure $S_0$ is the DAG without arcs at all and the local probability distributions are those given by the $n$ uni-dimensional marginal probabilities $p(X_i = x_i) = \frac{1}{r_i}$, $i = 1, \ldots, n$. This means that $M_0$ assigns the same probability to all individuals. The learning of $M_1$ is done by using the Algorithm B, while the rest of the models are learnt by means of local search. This local search start with the model obtained in the previous generation.

### 4.3 Penalized marginal likelihood $EBNA_{K2+pen}$

In this approach we propose to combine the Bayesian approach to calculate the marginal likelihood introduced by Cooper and Herskovits (1992) in addition with a penalizing term introduced to avoid a more complex Bayesian network. Thus, we propose to search for the Bayesian network structure $S$ that maximizes the expression:

$$\log p(D \mid S) - f(N) \dim(S). \quad (7)$$

Using the K2 metric for calculate $\log p(D \mid S)$, the expression we want to maximize is:

$$\log \left[ \prod_{i=1}^{n} \prod_{j=1}^{q_i} \frac{(r_i - 1)!}{(N_{ij} + r_i - 1)!} \prod_{k=1}^{r_i} N_{ijk}! \right] - f(N) \sum_{i=1}^{n} (r_i - 1) q_i. \quad (8)$$

Although we are aware of the implicit penalization of complex structures that the Bayesian approach involves, we understand that the addition of a explicit penalization term to the metric derived by Cooper and Herskovits (1992) might be of interest in this new framework.

To carry out such a purpose, we use a theorem proved by Etxeberria et al. (1997a) which bounds the number of parent nodes in the Bayesian network with the best score for each variable. The application of the theorem allows us to reduce the search space and, above all, to determine the maximum number of parents in the optimal learnt Bayesian network for each variable. The improvement of the approach we propose here is concerned with the fact that the bound for the cardinality of the parent set is obtained in an automatic way and can be different for each variable of the domain. Our approach is based on the following theorem:

**Theorem 1 (Etxeberria et al., 1997a)** *Let $X = (X_1, \ldots, X_n)$ an n-dimensional variable. Let $r_i$ be the number of values that an unidimensional variable $X_i$ can have, with $i = 1, \ldots, n$. Let D be a database with N cases over $\boldsymbol{X}$. Consider the variable $X_i$. Let $m, l \in \mathbf{N}$ where $l < r_i$ and $N = r_i m + l$. Let $pa \in \mathbf{N}$ where*

$$\prod_{j=1}^{pa+1} r'_j - \prod_{j=n-pa}^{n-1} r'_j$$

$$> \frac{1}{(r_i-1)f(N)} \log \left[ \frac{N!(r_i+l-1)!}{(N+r_i-1)!} \left( \frac{(2r_i-1)!}{(r_i-1)!} \right)^m \right] \quad (9)$$

*$r'_1, \ldots r'_{n-1}$ being $r_1, \ldots, r_{i-1}, r_{i+1}, \ldots, r_n$ sort in ascending order. Then, the variable $X_i$ will not have more than pa parent nodes in the network structure which maximizes:*

$$\log \left[ \prod_{i=1}^{n} \prod_{j=1}^{q_i} \frac{(r_i-1)!}{(N_{ij}+r_i-1)!} \prod_{k=1}^{r_i} N_{ijk}! \right] - f(N) \sum_{i=1}^{n} (r_i-1) q_i. \quad (10)$$

As an illustrative example of application of the previous theorem, let us consider a domain with $n = 20$ variables, where the set containing the number of different values for each variables is given by

$$\{r_1, \ldots, r_i, \ldots, r_{20}\} = \{\overbrace{3, \ldots, 3}^{7}, 4, \overbrace{3, \ldots, 3}^{5}, 4, \overbrace{3, \ldots, 3}^{5}, 4\} \quad (11)$$

Suppose that the database $D$ contains $N = 422$ cases and that we use the penalizing function known as the Akaike information criterion, that is $f(N) = 1$, (Akaike, 1974). Using theorem 1, we obtain the upper bound for the parent nodes of variable $X_8$ in the Bayesian network with the best score. Dividing $N(422)$ by $r_8(4)$ we obtain $m = 105$ and $l = 2$. On the other hand, the set $\{r'_1, \ldots, r'_i, \ldots, r_{19}\} = \{\overbrace{3, \ldots, 3}^{17}, 4, 4\}$. Calculating the right side of equation (9) we obtain: $\frac{1}{(r_8-1)f(N)} \log \left( \frac{N!(r_8+l-1)!}{(N+r_8-1)!} \left( \frac{(2r_8-1)!}{(r_8-1)!} \right)^m \right) = 231.0034$. The first natural number, $pa$, for which $\prod_{j=1}^{pa+1} r'_j - \prod_{j=n-pa}^{n-1} r'_j$ is bigger than 231.0034 is $pa = 5$, in fact $\prod_{j=1}^{5+1} r'_j - \prod_{j=n-5}^{n-1} r'_j = 3^6 - 3^3 4^2 = 297$. This means that $X_8$ will not have more than 5 parent nodes in the structure which maximizes $\log \left[ \prod_{i=1}^{n} \prod_{j=1}^{q_i} \frac{(r_i-1)!}{(N_{ij}+r_i-1)!} \prod_{k=1}^{r_i} N_{ijk}! \right] - f(N) \sum_{i=1}^{n} (r_i-1) q_i$.

The experimental results presented by Etxeberria et al. (1997a) show that the application of this score to the structure learning of Bayesian networks reduces the complexity without affecting the quality of the learnt models.

## 5 Experimental results

### 5.1 Introduction

In order to measure the performance of the new proposed algorithms, $EBNA_{PC}$, $EBNA_{BIC}$ and $EGNA_{K2+pen}$, we have carried out some experiments. We check empirically in four problems the validity of our approaches against two simple type of EDAs, the $UMDA$ and $MIMIC$ algorithms and against the Genetic Algorithm $(GA)$.

### 5.2 The problems

*OneMax problem*

This is an easy linear problem. It can be written mathematically as:

$$F_1 = F_{OneMax}(\boldsymbol{x}) = \sum_{i=1}^{n} x_i. \quad (12)$$

The objective is to maximize the function $F_1$ with $x_i \in \{0, 1\}$.

*Checkerboard problem*

In this problem, a $s \times s$ grid is given. Each point of the grid can take two values 0 or 1. The aim of the problem is to create a checkerboard pattern of 0's and 1's on the grid. Each location with a value of 1 should be surrounded in all four directions by a value of 0, and vice-versa. Only the four primary directions are considered in the evaluation. The evaluation is measured by counting the number of correct surrounding bits for the present value of each bit position for a $(s-2) \times (s-2)$ grid. In this manner, the corners are not included in the evaluation. The maximum value is $4(s-2)^2$. The individuals of this problem have dimension $n = s^2$.

If we consider the grid as a matrix $C = [c_{ij}]_{i,j=1,\ldots,s}$ and interpret $\delta(a, b)$ as the Kronecker's delta function, the checkerboard function can be written as follows:

$$F_2 = F_{Checkerboard}(C) = 4(s-2)^2 -$$

$$\sum_{i=2}^{n-1}\sum_{j=2}^{n-1} \delta(c_{ij}, c_{i-1j}) + \delta(c_{ij}, c_{i+1j}) + \delta(c_{ij}, c_{ij-1}) + \delta(c_{ij}, c_{ij+1}) \quad (13)$$

*SixPeaks problem*

This maximization problem can be defined mathematically as:

$$F_3 = F_{SixPeaks}(x, t) =$$
$$max\{tail(0, x), head(1, x), tail(1, x), head(0, x)\} + R(x, t) \quad (14)$$

where

$$tail(b, x) = \text{number of trailing b's in } x$$

$$head(b, x) = \text{number of leading b's in } x$$

$$R(x, t) = \begin{cases} n & if \quad tail(0, x) > t \text{ and } head(1, x) > t \text{ or} \\ & \quad tail(1, x) > t \text{ and } head(0, x) > t \\ 0 & otherwise. \end{cases}$$

In our experiments, the value of $t$ was set to 30% of $n$.

*EqualProducts problem*

Given a set $n$ real numbers $\{b_1, b_2, \ldots, b_n\}$, a subset of them is chosen. The objective is to minimize the difference between the products of the selected and unselected numbers. Mathematically:

$$F_4 = F_{EqualProducts}(x) = \left| \prod_{i=1}^{n} x_i b_i - \prod_{i=1}^{n} (1 - x_i) b_i \right|. \quad (15)$$

In our experiments the set of the $n$ real numbers was built sampling a uniform distribution in the interval $[0, 4]$. Thus, we do not know the optimum of this problem.

### 5.3 General considerations

In this section we will set up some of the parameters and decisions that are common to all EDA applications and those that are necessary in the GA approach. In particular the characteristics to define in EDA approaches are: size of the population, stopping condition, selection method and number of selected individuals.

The size of the population is the same in all the approaches. It was set up depending on the complexity of the problem. These values –chosen after carrying out some experiments– were assigned to 512, 1000, 1600 and 1600 respectively.

Three stopping conditions were taken into account. Firstly, the algorithm is stopped when a fixed number of function evaluations is reached. Secondly, the algorithm is stopped when the average value of the individuals in the population improves less than a fixed value. Thirdly the algorithm is stopped when it finds the optimum in case it is known. It does not matter what condition is first met, whenever one of them occurs the algorithm is stopped.

We use truncation selection as the selection method of choice, i.e. the best individuals are selected. The number of selected individuals was set up to half of the population.

Some of this information is summarized in Table 1.

Table 1: Characteristics of the problems used in the experiments

| Problem | $F_1$ | $F_2$ | $F_3$ | $F_4$ |
|---|---|---|---|---|
| Dimension | 128 | 100 | 50 | 50 |
| Max. Evaluat. | 100,000 | 100,000 | 300,000 | 300,000 |
| Type | Max. | Max. | Max. | Min. |
| Optimum | 128 | 256 | 84 | — |

A GA was implemented using the GENITOR software (Whitley and Kauth 1988). The classical mutation and one-point crossover operators together with rank-based selection were used.

We carried out 100 experiments for each function and algorithm.

### 5.4 Results

Table 2 summarizes the mean value of the results obtained for each algorithm in each problem before the algorithm stopped. For a more detailed analysis the reader can consult Larrañaga et al. (1999b).

Table 2: Mean values reached by the algorithms in the problems

| Problem | $F_1$ | $F_2$ | $F_3$ | $F_4$ |
|---|---|---|---|---|
| $UMDA$ | 128 | 241.85 | 62.1 | 37.34 |
| $MIMIC$ | 128 | 243.34 | 57.86 | 38.43 |
| $EBNA_{PC}$ | 128 | 243 | 51.55 | 5,317.9 |
| $EBNA_{BIC}$ | 128 | 254.21 | 71.9 | 26.43 |
| $EBNA_{K2+pen}$ | 128 | 254.92 | 84 | 33.66 |
| $GA$ | 125.75 | 246.11 | 83.11 | 93.00 |

A deeper analysis of the results has been carried out by means of statistical tests. We have performed a Kruskal-Wallis test ($\alpha = 0.05$) related to the objective value reached for the algorithms. In all the problems, the results showed that significative differences between the algorithms exist.

Moreover, five more Kruskal-Wallis test were made for each problem. For this task, the algorithms were ranked with respect to the mean function value. Each algorithm was tested with its next algorithm in the rank order. For instance, in the case of the Checkerboard problem, the algorithms are ranked as follows: $EBNA_{K2+pen}$, $EBNA_{BIC}$, $GA$, $MIMIC$, $EBNA_{PC}$ and $UMDA$. For this case, we tested, $EBNA_{K2+pen}$ against $EBNA_{BIC}$, later $EBNA_{BIC}$ against $GA$, and so on.

Table 3 summarizes the tests results. The entries of the table show the existence or absence of significative differences. The same number for two consecutive algorithms (in the rank order) means that there is no significative difference. For example, in the Checkerboard problem, because $EBNA_{K2+pen}$ and $EBNA_{BIC}$ have different numbers, this means that a significative difference between both algorithms exists. However, $EBNA_{BIC}$, $GA$, $MIMIC$ and $EBNA_{PC}$ were assigned the same number, which means that there is no significative difference between $EBNA_{BIC}$ and $GA$, between $GA$ and $MIMIC$, and between $MIMIC$ and $EBNA_{PC}$. The particular numbers in each entry represent the number of algorithms that obtained better results (for the test carried out), where 1 means the best algorithm.

Table 3: Results of the Kruskal-Wallis tests between pair of algorithms

| Problem | $F_1$ | $F_2$ | $F_3$ | $F_4$ |
| --- | --- | --- | --- | --- |
| $UMDA$ | 1 | 6 | 4 | 1 |
| $MIMIC$ | 1 | 2 | 4 | 1 |
| $EBNA_{PC}$ | 1 | 2 | 6 | 6 |
| $EBNA_{BIC}$ | 1 | 2 | 3 | 1 |
| $EBNA_{K2+pen}$ | 1 | 1 | 1 | 1 |
| $GA$ | 6 | 2 | 2 | 5 |

As a result of the experiments, we conclude that $EBNA_{K2+pen}$ is the algorithm that returned the best results. $EBNA_{BIC}$ obtains results that are near those obtained by $EBNA_{K2+pen}$. On the contrary, the worst results were obtained with $EBNA_{PC}$ and $GA$.

## 6  Conclusions and futher work

In this paper, we have introduced new approaches for combinatorial optimization based on the structure learning of Bayesian networks from different perspectives –detecting independencies and by penalized maximum likelihood and marginal likelihood. Also empirical evidence of the behavior of the proposed algorithms has been shown.

There remain some open questions in this new framework for optimization. Among them, the determination of an appropiate sample size, the development of techniques able to deal with optimization under constraints, the parallelization of the diferents tasks, and the study of the behavoir of EDA from a mathematical point of view.

### Acknowledgement

This work was partially supported by the Spanish *Comisión Interministerial de Ciencia y Tecnología* (CICYT) under project TIC97-1135-C04-03, by the University of the Basque Country under project UPV 140.226-EB131/99, and by the *Departamento de Educación, Universidades e Investigación* under grant PI98/74. We thank E. Bengoetxea and the anonymous reviewers for useful suggestions.

## References

[1] H. Akaike (1978). New Look at the Statistical Model Identification. *IEEE Transactions on Automatic Control* **19**, 6, 716–723.

[2] S. Baluja (1994). Population-Based Incremental Learning: A Method for Integrating Genetic Search Based Function Optimization and Competitive Learning. *Carnegie Mellon Report. CMU-CS-94-163*.

[3] S. Baluja and S. Davies (1997). Using Optimal Dependency-Trees for Combinatorial Optimization: Learning the Structure of the Search Space. *Carnegie Mellon Report. CMU-CS-97-107*.

[4] W. Buntine (1991). Theory refinement in Bayesian networks. *Proceedings of the Seventh Conference on Uncertainty in Artificial Intelligence*, 52–60.

[5] W. Buntine (1996). A Guide to the Literature on Learning Probabilistic Networks from Data. *IEEE Transactions on Knowledge and Data Engineering* **8**: 2, 195–210.

[6] E. Castillo, J. M. Gutierrez, and A. S. Hadi (1997). *Expert Systems and Probabilistic Network Models*. Springer-Verlag.

[7] D. M. Chickering, D. Geiger, and D. Heckerman (1994). Learning Bayesian networks is NP-hard. *Technical Report MSR-TR-94-17*. Microsoft Research.

[8] D. M. Chickering, D. Geiger, and D. Heckerman (1995). Learning Bayesian Networks: Search Methods and Experimental Results. *Preliminary Papers of the Fifth International Workshop on Artificial Intelligence and Statistics*, 112–128.

[9] G. F. Cooper and E. A. Herskovits (1992). A Bayesian method for the induction of probabilistic networks from data. *Machine Learning* **9**, 309–347.

[10] R. G. Cowell, A. P. Dawid, S. L. Lauritzen and D. J. Spiegelhalter (1999). *Probabilistic Networks and Expert Systems*. Springer-Verlag.

[11] J. S. De Bonet, C. L. Isbell, and P. Viola (1997). MIMIC: Finding Optima by Estimating Probability Densities, *Advances in Neural Information Processing Systems, Vol. 9*.

[12] R. Etxeberria, P. Larrañaga, and J. M. Picaza (1997a). Reducing Bayesian networks complexity while learning from data. *Proceedings of Causal Models and Statistical Learning*, 151–168.

[13] R. Etxeberria, P. Larrañaga, and J. M. Picaza (1997b). Analysis of the behaviour of the genetic algorithms when searching Bayesian networks from data. *Pattern Recognition Letters* **18: 11–13**, 1269-1273.

[14] R. Etxeberria and P. Larrañaga (1999). Global optimization with Bayesian networks, *II Symposium on Artificial Intelligence. CIMAF99. Special Session on Distributions and Evolutionary Optimization*, 332–339.

[15] G. Harik (1999). Linkage learning in via Probabilistic Modeling in the ECGA. *IlliGAL Technical Report, No. 99010*.

[16] G. Harik, F. G. Lobo, and D. E. Goldberg (1998). The Compact Genetic Algorithm. *Proceedings of the IEEE Conference on Evolutionary Computation*, 523–528.

[17] D. Heckerman (1995). A Tutorial on Learning with Bayesian Networks. *Technical Report MSR-TR-95-06. Microsoft Research*.

[18] M. Henrion (1988). Propagating uncertainty in Bayesian networks by probabilistic logic sampling. *Uncertainty in Artificial Intelligence* **2**, 149-163.

[19] I. Inza, P. Larrañaga, R. Etxeberria and B. Sierra (1999). Feature Subset Selection by estimation and simulation of Bayesian networks. *Technical Report, University of the Basque Country, KZAA-IK-99-02*.

[20] F. V. Jensen (1996). *An introduction to Bayesian networks*. University College of London.

[21] P. Larrañaga, M. Poza, Y. Yurramendi, R. H. Murga, and C. M. H. Kuijpers (1996a). Structure learning of Bayesian networks by genetic algorithms: A performance analysis of control parameters. *IEEE Transactions on Pattern Analysis and Machine Intelligence* **18 (9)**, 912–926.

[22] P. Larrañaga, C. M. H. Kuijpers, R. H. Murga, and Y. Yurramendi (1996b). Searching for the best ordering in the structure learning of Bayesian networks, *IEEE Transactions on Systems, Man and Cybernetics* **26 (4)**, 487–493.

[23] P. Larrañaga, R. Etxeberria, J. A. Lozano, B. Sierra, I. Inza, and J. M. Peña (1999a). A review of the cooperation between evolutionary computation and probabilistic graphical models. *Second Symposium on Artificial Intelligence. Adaptive Systems. CIMAF 99*, 314-324.

[24] P. Larrañaga, R. Etxeberria, J. A. Lozano, and J. M. Peña (1999b). Optimization by learning and simulation of Bayesian and Gaussian networks. *Technical Report, University of the Basque Country, KZAA-IK-99-04*.

[25] H. Mühlenbein (1998). The Equation for Response to Selection and its Use for Prediction, *Evolutionary Computation* **5**, 303–346.

[26] H. Mühlenbein and G. Paaß(1996). From Recombination of Genes to the Estimation of Distributions I. Binary Parameters. *Lecture Notes in Computer Science 1411: Parallel Problem Solving from Nature–PPSN IV*, 178–187.

[27] H. Mühlenbein, T. Mahnig, and A. Ochoa (1999). Schemata, Distributions and Graphical Models in Evolutionary Computation. *Journal of Heuristics* **5**, 215–247.

[28] H. Mühlenbein and T. Mahnig (1999). FDA - A scalable evolutionary algorithm for the optimization of additively decomposed functions. *Evolutionary Computation* **7**(4), 353–376.

[29] J. W. Myers, K. B. Laskey and T. Levitt (1999). Learning Bayesian Networks from Incomplete data with Stochastic Search Algorithms. *Proceedings of the Fifteenth Conference on Uncertainty in Artificial Intelligence*, 476–485.

[30] J. Pearl (1988). *Probabilistic Reasoning in Intelligent Systems*. Morgan Kaufmann.

[31] M. Pelikan and H. Mühlenbein (1999). The Bivariate Marginal Distribution Algorithm, *Advances in Soft Computing-Engineering Design and Manufacturing*. Springer-Verlag, 521–535.

[32] M. Pelikan, D. E. Goldberg and E. Cantú-Paz (1999). BOA: The Bayesian optimization algorithm. *Proceedings of the Genetic and Evolutionary Computation Conference GECCO-99*, Vol. 1, Morgan Kaufmann Publishers, 525–532.

[33] G. Schwarz (1978). Estimating the dimension of a model. *Annals of Statistics* **7(2)**, 461–464.

[34] P. Spirtes, C. Glymour, and R. Scheines (1991). An algorithm for fast recovery of sparse causal graphs. *Social Science Computing Reviews* **9**, 62–72.

[35] G. Syswerda (1993). Simulated crossover in genetic algorithms. *Foundations of Genetic Algorithms* **2**, 239–255, Morgan Kaufmann.

[36] D. Whitley and J. Kauth (1988). GENITOR: A different genetic algorithm, *Proceedings of the Rocky Mountain Conference on Artificial Intelligence*, Vol. 2, 118–130.